\documentclass[10pt, conference, letterpaper]{IEEEtran}
\IEEEoverridecommandlockouts

\usepackage[labelformat=simple]{subcaption}

\usepackage{cite}
\usepackage{graphicx}
\usepackage{textcomp}
\usepackage{xcolor}

\usepackage{url}
\usepackage[colorlinks,citecolor=blue,urlcolor=blue,linkcolor=blue,bookmarks=false,hypertexnames=true]{hyperref}
\usepackage{multirow}

\usepackage{xcolor, xspace}

\usepackage{amsmath,amssymb,amsfonts}
\usepackage{algorithmic}
\usepackage{graphicx}
\usepackage{textcomp}
\usepackage{xcolor}
\newcommand{\algo}{\texttt{GWTF}\xspace}

\newif\ifshortversion

\newcommand{\hide}[1]{\ifshortversion\else{#1}\fi}

  \author{\IEEEauthorblockN{Nikolay Blagoev\IEEEauthorrefmark{1}\IEEEauthorrefmark{2}, Bart Cox\IEEEauthorrefmark{3}, Jérémie Decouchant\IEEEauthorrefmark{3}, Lydia Y. Chen\IEEEauthorrefmark{2}}
  \IEEEauthorblockA{
    \textit{\IEEEauthorrefmark{1}Worker Thread, \IEEEauthorrefmark{2}Université de Neuchâtel, \IEEEauthorrefmark{3}Delft University of Technology}\\
    nickyblagoev@gmail.com, b.a.cox@tudelft.nl, j.decouchant@tudelft.nl, yiyu.chen@unine.ch}
  }
    
\begin{document}

\title{Go With The Flow: Churn-Tolerant Decentralized Training of Large Language Models}

\maketitle

\begin{abstract}
Motivated by the emergence of large language models (LLMs) and the importance of democratizing their training, we propose \algo, the first crash tolerant practical decentralized training framework for LLMs. Differently from existing distributed and federated training frameworks, \algo{} enables the efficient collaborative training of a LLM on heterogeneous clients that volunteer their resources. In addition, \algo{} addresses node churn, i.e., clients joining or leaving the system at any time, and network instabilities, i.e., network links becoming unstable or unreliable. The core of \algo{} is a novel decentralized flow algorithm that finds the most effective routing that maximizes the number of microbatches trained with the lowest possible delay. We extensively evaluate \algo{} on GPT-like and LLaMa-like models and compare it against the prior art. Our results indicate that \algo{} reduces the training time by up to 45\%  in realistic and challenging scenarios that involve heterogeneous client nodes distributed over 10 different geographic locations with a high node churn rate. 
\end{abstract}

\begin{IEEEkeywords}
Large Language Model, Decentralized learning, Crash-tolerance, Flow optimization.
\end{IEEEkeywords}

\section{Introduction}

 Deep transformer-based architectures have recently enabled unprecedented performance on tasks such as next word prediction~\cite{gpt} and image recognition~\cite{VIT} thanks to the growing corpora of available data and the increasing size of Large Language Models (LLMs)~\cite{llama,dectrain}. However, models are now too large to fit and be efficiently trained on a single GPU. 
 Parallel training techniques, such as Pipeline Parallelism (PP) and Data Parallelism (DP), therefore need to be used to efficiently  train large models on clusters of nodes. However, renting cloud resources or using private computer clusters to train models can easily cost more than tens of thousands of dollars~\cite{dectrain}, even for smaller models. Developing and training LLMs this way is therefore out of the reach of the public. 

To democratize access to language models, a promising alternative to using private clusters is volunteer computing~\cite{foldinghome}, i.e., using spare computing resources from independent participants around the world.
The potential of volunteer computing for the development of large deep networks has been acknowledged in recent publications~\cite{dectrain,swarm}.
To realize this potential, the decentralized training of language models requires facing fault-tolerance and performance challenges. First, recovery mechanisms are necessary to tolerate participants freely joining or leaving the system at any time, which also implies that participants can only have partial knowledge of the global system membership. Second, reactive mechanisms are required to adapt to networks becoming unstable or unreliable, and to participants dedicating fluctuating computing resources over time.  

Recently, SWARM, a pioneering work by Ryabinin et al.~\cite{swarm}, has taken a step towards addressing these issues. However, SWARM fails to fully utilize its resources. First, it fails to properly recover from crashes in the backwards pass. Second, it uses a simple greedy approach during routing and therefore does not aim at optimizing training time.
Finally, SWARM assumes that all nodes have the same amount of memory, which in a highly decentralized setting is an unrealistic expectation to hold. Our work addresses these limitations. 

We present Go With The Flow (\algo), a decentralized and efficient LLM training framework. We model the routing of microbatches between heterogeneous nodes in the forward and backward pass of stochastic gradient descent as a minimum cost flow problem and optimize its execution on geographically dispersed nodes in a decentralized manner. \algo{} minimizes the training time and maximizes its throughput by continuously addressing the bottleneck stages of LLM training. We evaluate \algo{} on the training of GPT-like and LLama-like models of 300\,M and 7\,B parameters against SWARM and show an up to 40\% training time reduction while maximally utilizing the available clients.

This paper makes the following \textbf{contributions}: \\
$\bullet$ We propose and implement \algo{}, an efficient churn-tolerant decentralized training framework for LLMs. \\
$\bullet$ \algo{} utilizes a novel fully decentralized minimum cost flow algorithm, empowering  individuals to contribute their heterogeneous and volatile resources to LLM training. \\
$\bullet$ \algo{} handles efficiently node churn of LLM during both the forward and the backward passes of stochastic gradient descent, by instantaneously rerouting to available nodes thanks to its decentralized nature.  \\
$\bullet$ \algo{} minimizes the training time and maximizes the throughput of training LLMs in heterogeneous crash-prone environments, but also optimally utilizes available resources, without sacrificing convergence. 

\hide{
This paper is organized as follows. 
Section~\ref{sec:background} provides necessary background knowledge and discusses the related work. 
Section~\ref{sec:sysmodel} states our system model and our objectives. 
Section~\ref{sec:gwtf} presents an overview of \algo{}, and Section~\ref{sec:details} details the main algorithms of \algo{} that optimize flows in a decentralized manner, tolerate crashes and synchronize training and aggregations.
Section~\ref{sec:eval} presents our performance evaluation. 
Finally, Section~\ref{sec:conclusion} concludes this paper. 
}
\section{Background \& Related Work}
\label{sec:background}

\begin{table}
  \caption{Notations}
  \small
  \label{notation-table}
  \centering
  
  \begin{tabular}{ll}
    \hline
    Name     & Description         \\
    \hline
    \textbf{S}     & A single stage - a set of nodes\\
    \textbf{N}     & All nodes/peers           \\
    \(T\)& Temperature     \\
    \(\alpha\)& Cooling factor    \\
    \(d(i,j)\) & Cost between nodes $i$ and $j$\\
    \(f(i,j)\) & Flow between nodes $i$ and $j$\\
    \({cost}_f\) & Cost of a single flow $f$ \\
    \({cap}_i\) & Memory capacity  of node $i$ \\  
    \(U(x,y)\) & Uniform random number between $x$ and $y$ \\ 
    \(c_i\) & Computation time/cost of node \(i\) \\
    \(\lambda_{i,j}\) & Network latency between nodes \(i\) and \(j\) \\
    \(\beta_{i,j}\) & Network bandwidth between nodes \(i\) and \(j\) \\
    \hline
  \end{tabular}

\end{table}

\begin{figure*}
  \centering
  \includegraphics[width = 0.85\textwidth]{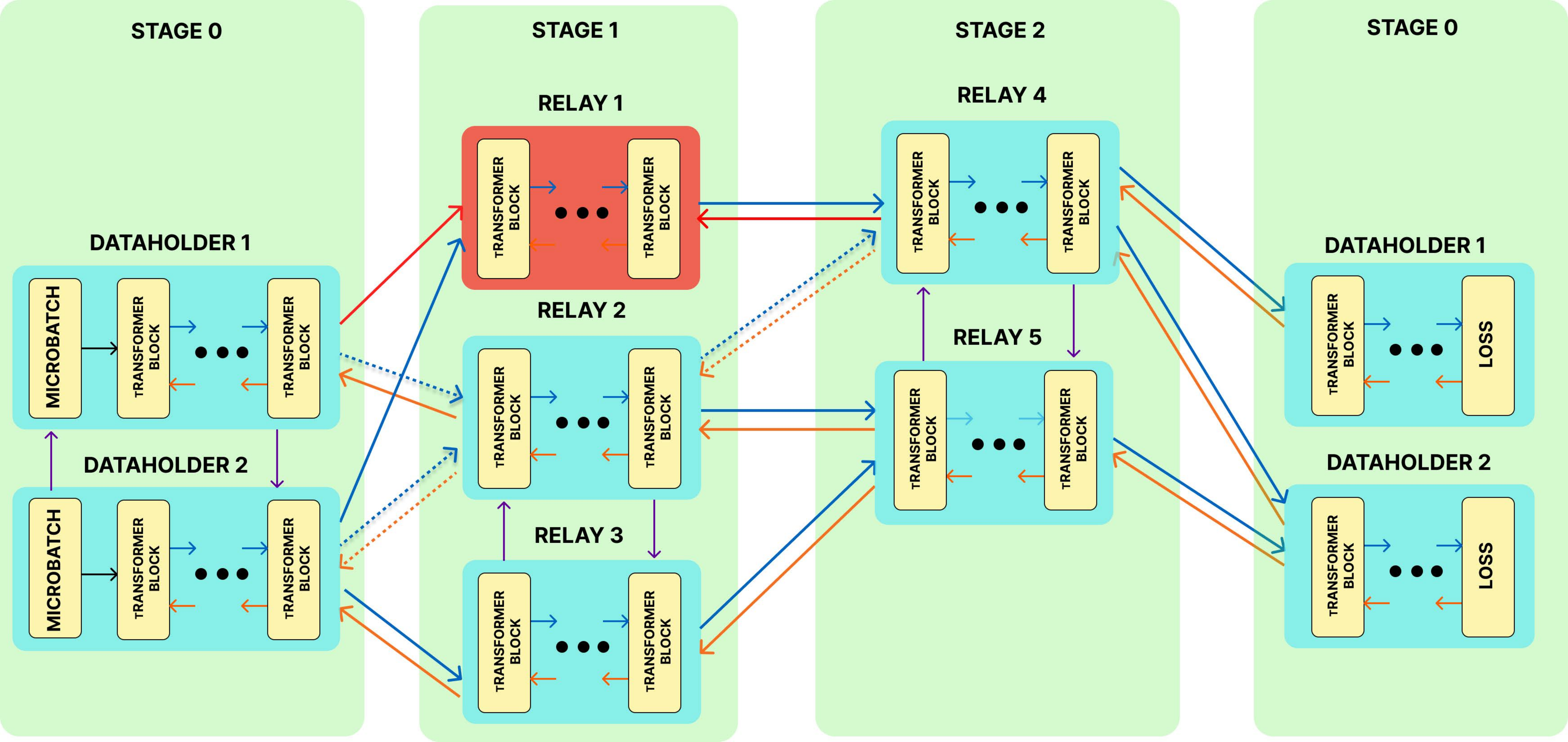}
  \caption{
  Crash-recovery during decentralized training of an LLM with \algo. The plain blue arrows between nodes indicate the forward passes, the plain orange arrows indicate the backward passes, the plain purple ones indicate model aggregations, and the red arrows indicate some network transmissions that fail because relay 1 crashes (shown in red). Dashed lines indicate the newly formed microbatch exchanges that are generated to recover from the crash of relay node 1.}
  \label{fig:overview}
\end{figure*}

\hide{
\textbf{Large Language Models.} 
Large Language Models are typically constructed as an embedding layer and a series of transformer blocks~\cite{llama, gpt,palm,megatronlm}. Embedding layers simply translate the input which is in some high dimension, to a lower dimensional output, which makes learning easier for the neural network. A transformer consists of multi-head self-attention, normalization, and a feed forward network \cite{transformer}.
}

\textbf{Pipeline Parallelism.}
Consecutive layers (i.e., transformer blocks) of an LLM are grouped into stages that are distributed over several devices, forming a pipeline. The first stage retrieves the (tokenized) data and embeds it, while the last stage evaluates the loss function. 
The first and last stages of the pipeline are colocated on the same device (the data node) because they require the training data. All stages hold one or several  contiguous transformer blocks. Together, they behave as a single model. To train a model three computation phases are executed: (i) the forward pass (from the first stage to the last stage) to compute the loss; (ii) the backward pass (from the last stage back to the first stage) to compute the gradients per parameter; and (iii) the update phase, where model weights are updated based on the gradients.

\textbf{Microbatches.} 
Each batch is further split into microbatches~\cite{gpipe} so that 
multiple microbatches can be concurrently processed on different devices. Following the processing of all the microbatches of a batch, nodes need to update their model parameters based on the average of the microbatches' gradient updates they have stored until this point. An iteration is thus defined as the processing of one batch, and its duration is defined as the elapsed time between the end of two consecutive update phases (measured from the viewpoint of the slowest data node). 

\textbf{Data Parallelism.} 
\hide{
A larger batch size allows SGD to converge faster~\cite{batchsgd}. However, in practice, the amount of memory available on devices limits the maximum batch size that can be used. 
Data parallelism is another way to circumvent this limitation.
}
Data parallelism consists in distributing the processing of microbatches for a given stage over several devices that each host an identical model, which accelerates convergence~\cite{batchsgd}. These devices concurrently perform iterations on their microbatches. At the end of each iteration, and before the update phase, these devices aggregate their collective work by exchanging their stored gradients (aggregation phase). Later on, they perform the update phase based on the average of all gradients.  


\textbf{Decentralized Pipeline Parallelism.} 
Few works considered the decentralized training of LLMs.
\cite{dectrain} compute a communication-optimal arrangement of nodes in pipelines via a centralized and computationally-expensive genetic algorithm, but ignore churn.  
In SWARM~\cite{swarm}, nodes independently route a micro-batch through stages. A node sends its activations (outputs of the last layer of its transformer model) to a next stage's node. If a node fails to process a microbatch before some predefined time, the previous node sends its activations to a different node. However, SWARM 
does not address crashes that may occur during the backward pass. 
SWARM's timeout-resend strategy requires a complete pipeline recomputation, wasting significant resources. Additionally, SWARM nodes employ a greedy procedure to select their next stage successor, which does not guarantee optimal pipeline duration or takes a device's memory constraints into account. 

\hide{
\textbf{Gossip Learning.} Previous works on decentralized training mainly focused on exploiting data parallelism, and predominantly relied on gossip-based  communications~\cite{gossipnew,asymgossip,epidemiclearning}. In those approaches, during the parameter sharing step, nodes send their model weights to a subset (of size \(k\) - the group size) of known peers. After receiving all the weights they expect during an iteration, including their own, nodes compute their average and use it in the following iteration. Gossip learning  can naturally tolerate device crashes, dynamic communication graphs, heterogeneous networks (by preferring faster links), and only requires nodes to maintain partial membership knowledge.
}

\hide{
\begin{figure}
  \centering
  \includegraphics[width =\columnwidth]{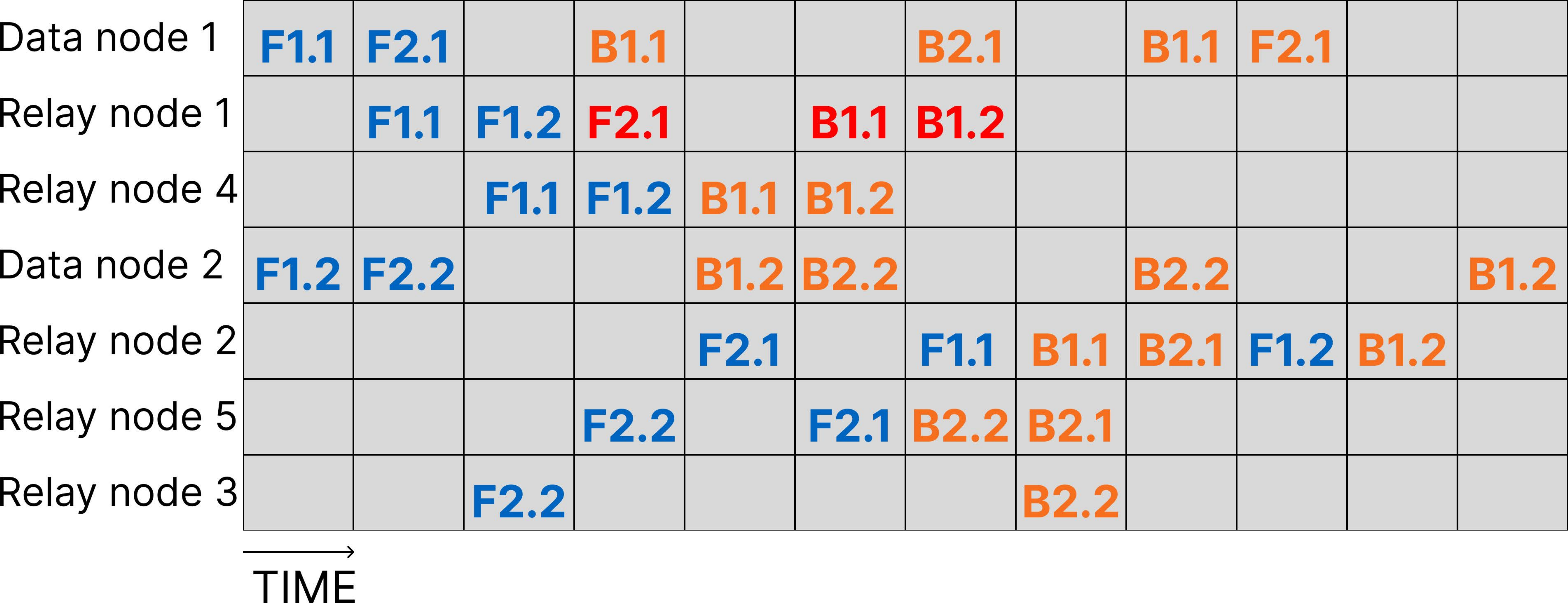}
  \caption{Execution scenario of the decentralized training of an LLM. Notations $Fx,y$ and $Bx,y$ to indicate a forward pass, respectively a backward pass, using microbatch $x$ generated by data node $y$.  
  Blue is used for a forward pass, orange for a backward pass, and red indicates a task that is not executed because of a node crash.  }
  \label{fig:time-execution}
\end{figure}
}

 \section{System Model and Objectives}
\label{sec:sysmodel}

\textbf{System model.} 
Nodes independently contribute resources for limited periods and are geographically distributed with individual memory and communication constraints.  Each has a partial view of the system and communicates with known peers over authenticated, unreliable, and heterogeneous links. The network is partially synchronous, i.e., there are periods of time where the network latency is bounded.
We model a node $i$ with a given memory capacity with a maximum number $cap_i$ of microbatches it can process at a time, and denote by $d_{i,j}$ the communication delay between a node pair ($i$, $j$). 
Nodes act as data nodes, relay nodes, or both. The former holds training data, whereas the latter contribute to the training of the LLM. 

\textbf{Node churn.}
Nodes can join, leave, or crash at any time, including during forward or backward passes. We focus on the crashing case, which we resolve through timeouts. A node crashing during a forward pass slows down the construction of the forward pass. A node crashing during a backward pass requires additional synchronization between nodes across different stages in order to recover the pipeline by replacing the missing nodes. 

\textbf{Objectives.}
We aim to train LLMs efficiently under churn, heterogeneity, and unstable networks. Prior work~\cite{dectrain,sdpipe} assume stable and homogeneous conditions, which do not usually hold in decentralized settings, and crashes can halt training. 
SWARM~\cite{swarm}  addresses forward-pass faults via rerouting but still requires full pipeline recomputation on backward-pass failures, doubling training time.

Fig.~\ref{fig:overview} illustrates the training of an LLM with \algo{} using 3 stages, 2 data nodes that hold the training data, and 5 relay nodes that execute the intermediate training steps for two batches. Relay node 1 crashes when processing the microbatch of the first data node, and does not execute a forward pass from the second data holder, or a backwards pass for the first data node, which are redirected to relay node 2.
\hide{
Fig.~\ref{fig:time-execution} illustrates the corresponding execution scenario.
}

\section{\algo{} in a Nutshell}
\label{sec:gwtf}

\hide{This section highlights the cost minimization objective of \algo{} and its key design components.}

\textbf{Flow and Cost.} 
\algo{} aims at minimizing the distributed training cost of an LLM, which we define in the following manner. We model the average delay, termed cost, of exchanging activations between two nodes $i$ and $j$ using the sum of its computation times and communication delays. We term this "flow" - an abstraction of the end-to-end processing of a microbatch through the stages in the system. In Figure~\ref{fig:overview} we can see a single flow going from dataholder 2, to relay 3, to relay 5, back to dataholder 2. We use this notion of flow instead of the traditional pipeline parallelism, allowing nodes to communicate with multiple other nodes in the previous and subsequent pipeline. Following Yuan et al.~\cite{dectrain}, the communication cost between nodes $i$ and $j$ is composed of network latency (\(\lambda_{i,j}\)) and a transmission delay that is computed as the amount of transferred data ($size$) divided by the network bandwidth (\(\frac{size}{\beta_{i,j}}\)).
We assume that links are not necessarily symmetric, i.e., that $\lambda_{i,j}$ and $\beta_{i,j}$ are not necessarily respectively equal to $\lambda_{j,i}$ and $\beta_{j,i}$. However, as every link is used twice during training (once from \(i\) to \(j\) during the forward pass and once from \(j\) to \(i\) during the backward pass), we can model the latency on the link as the average of the two delays. The final communication cost between two nodes $i$ and $j$ in a given phase is then \(\frac{\lambda_{i,j}+\lambda_{j,i}}{2} + \frac{2 \cdot size}{\beta_{i,j} + \beta_{j,i}}\). We denote the computation cost of nodes $i$ and $j$ respectively by $c_{i}$ and $c_j$. Computation costs capture the time it takes a node to process a microbatch during the forward or the backward pass. The final cost $d_{i,j}$ of a microbatch flow between nodes $i$ and $j$ is therefore defined by the following equation:

\begin{equation}
\label{eq:t}
d_{i,j}=\frac{c_i + c_j}{2} + \frac{\lambda_{i,j} + \lambda_{j,i}}{2} +  \frac{2 \cdot size}{\beta_{i,j} + \beta_{j,i}}
\end{equation}

\algo{} aims at processing the largest number of microbatches at the lowest global cost, and uses a novel decentralized algorithm to route microbatches through the nodes. 

\textbf{Key Design Components.}
During each iteration, data nodes first send out the embedded microbatches to the relay nodes in the first stage. Each microbatch is associated with a unique id, the data node of its origin, and a path list, which holds the nodes it has traversed during its execution.

A relay node executes up to 5 different sub-routines when participating in the system: (1) forming of a pipeline with the flow algorithm; (2) joining the system; (3) resending in case of a node crash during forward pass; (4) recovering the pipeline in case of backward failure or lost batches; and (5) aggregating the models received from other nodes that belong in the same stage. 
The first four procedures run in parallel to the actual processing of forward and backward passes (the training phase). The last procedure is the model update phase. 

The decentralized flow cost minimization algorithm, 
reduces the cost of the forward and backward pass for each microbatch. It aims at sending a maximum number of batches at the smallest cost according to Equation~\ref{eq:t}. Nodes with different capacities route microbatches in a cost-effective manner, independently and without global knowledge.

Nodes are assigned to stages when they join the system. 
\algo{} aims to increase the throughput of the stage with the minimum capacity, as that stage puts a bottleneck on the current throughput in an iteration. Nodes discover other peers in the system through a Distributed Hash Table (DHT)~\cite{DHT}. An elected leader from the data nodes periodically adds new nodes, prioritizing nodes with higher capacity to join the stage with the smallest capacity, thus expanding the bottleneck of the system. Note that the leader can be elected in a robust way~\cite{garcia1982elections,ongaro2014search}.

\algo handles nodes leaving due to unavailability or crashes, 
which is more challenging than handling new joiners.
Crashes occurring during a forward pass are resolved by resending to another peer in the next stage, according to the new flow found by (Sec. \ref{routing}. During a fault in a backwards pass the microbatch list is used to replace the crashed nodes and restore the pipeline with a minimal amount of computations. More precisely, to amend a broken flow, \algo{} searches for the last alive node before a crash to quickly repair the flow to the first alive node after the crash, via nodes that have spare computational resources in between.

\section{\algo Details} 
\label{sec:details}

\hide{
This section first provides the details of \algo{}'s decentralized flow optimization, which aims at maximizing the training throughput of an LLM on nodes of different capacity at a minimal cost. Then, it discusses how \algo{} inserts new nodes into existing pipelines, and how pipelines are dynamically optimized. Finally, it explains how nodes crashing or leaving the system are tolerated.   
}

\subsection{Decentralized flow optimization}
\label{routing}

In parallel to the training and node addition, \algo constructs flows (abstract pipelines) for a microbatch starting from a data node and sequentially determining the relay nodes for each stage. To build a flow, the availability of known nodes and their memory constraints are considered. We assume that the memory requirement of each stage per microbatch is known through offline profiling, which can be done by the data nodes themselves and subsequently announced to all nodes. Using this information, nodes can determine their capacity based on their known memory.

The objective of \algo{} is to complete the entire flow of all microbatches at the minimum cost, namely the sum of \(d_{i,j}\) from Equation~\ref{eq:t} along the path of the flow.
This optimization problem is closely related to the problem of minimizing cost and maximizing flow in a multiple-source multiple-sink graph. The sources and the sinks are both the data nodes, as microbatches travel back to their origin for loss computation. Flow on a given edge/link equates to number of microbatches transferred between the two nodes connected by that link. A unit of flow is thus a microbatch. In such a problem the goal is to minimize the sum of costs of all flows within the graph: 
\begin{equation}
\label{eq:flwcst}
min \; \left( \sum_{ \forall i,j \in \textbf{N}} \; f(i,j) \cdot d(i, j)\right)
\end{equation}
where \(f(i,j)\) is the flow between two nodes on the link that connects the,. The problem assumes a linear model for cost to flow increase, hence the inner product of \( f(i,j) \cdot d(i, j)\) expressing the cost of some amount of flow along some edge.
The classical algorithm~\cite{outofkilter} one can use to solve our optimization problem relies on global information, which is not available in decentralized settings.  Additionally, unlike in the standard definition which requires that any flow from a source be delivered to any sink, \algo{}  has to deliver a flow from a source back to itself.

\par To this end we design a novel distributed algorithm that leverages only local knowledge and differentiates between flow from different sources. However, the objective function of Equation~\ref{eq:flwcst} requires greater synchronization between nodes  and converges slowly, as nodes need global knowledge of the cost of all flows. Fortunately we can solve a much easier problem with only local knowledge for each node by minimizing the cost of the maximal flow between two nodes:
\(
\mathrm{min} \; \left( \max_{ \forall i,j \in \textbf{N}} \; f(i,j) \cdot d(i, j)\right)
\).
A similar function has also been used in a previous work~\cite{dectrain}, where their goal is to minimize the maximum communication cost between two subsequent nodes in a pipeline. 

\subsection{Inserting Joining Nodes}
\label{sec:node-addition}

\hide{
We design a two-way procedure to assign a new node to the stage that is identified to be the bottleneck.
}
A data node is elected, which we call the leader, and is in charge of ranking the stages based on their utilized ratio, which is computed as the number of flows that go through them divided by their total available capacity. Stages with a higher utilization are given a higher priority to incorporate incoming nodes. Periodically, a flooding algorithm is used to find the utilization of each stage. This is initiated by the leader, who sends a query to all known peers in the following stage. Whenever a node receives a query from the previous stage, it adds its capacity and utilization to the message, and forwards it to known peers in the subsequent stage.

A joining node discovers other peers in the system and the leader's identity through a Distributed Hash Table (DHT)~\cite{DHT}. 
Upon joining the system, new nodes (candidates) send their capacity to the elected leader. The leader, periodically, selects the candidates with highest capacities and adds them to the most utilized stages. The candidate with the highest capacity will be added to the stage with highest utilization, the second highest to the stage with second highest, and so on. 

\hide{
    \begin{figure}[t]
      \centering
      \includegraphics[width=\columnwidth]{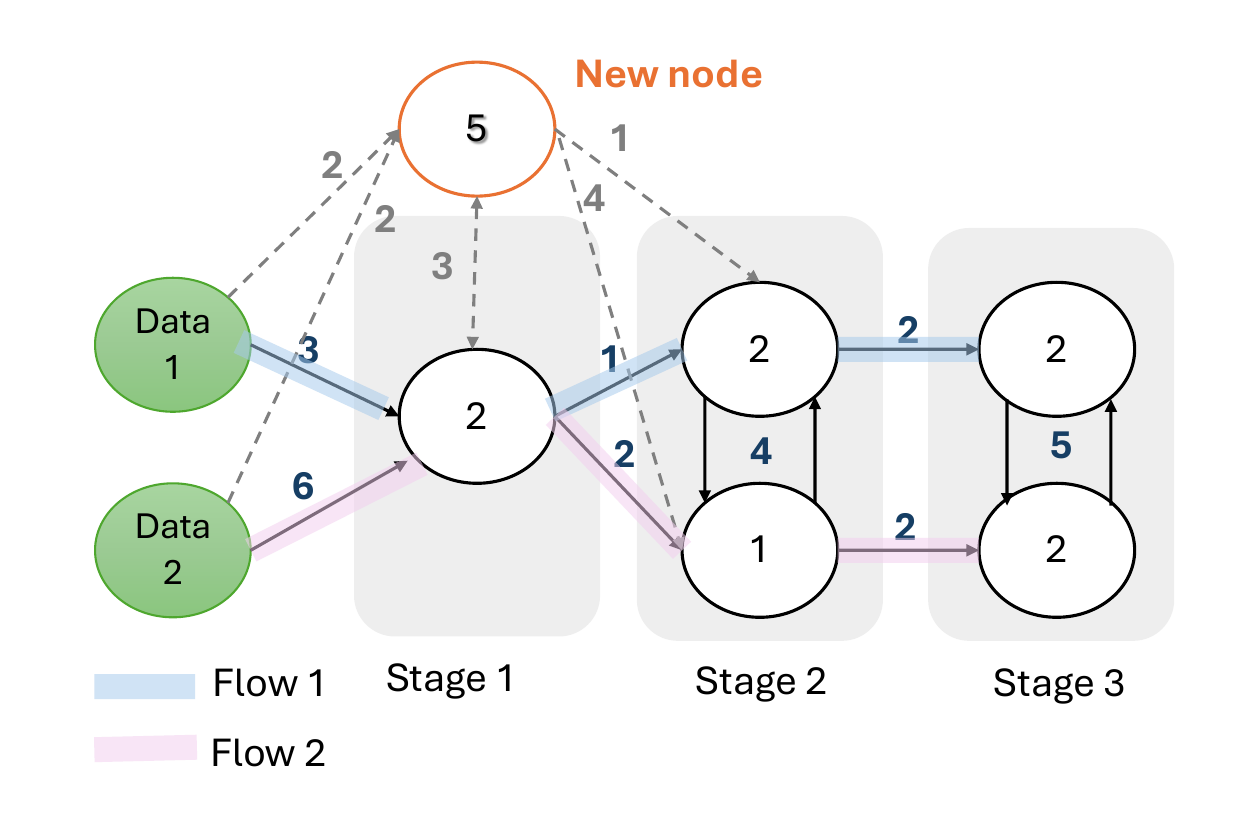}
      \caption{A joining node being added to stage 1 in a system that includes two data nodes and three relay nodes.}
      \label{fig:joining}
    \end{figure}
    Figure~\ref{fig:joining} illustrates \algo{}'s node joining procedure.
    In this example, there are 2 dataholder nodes (on the left, in green), each sending a single microbatch flow traversing 4 stages. Each node is indicated by a circle whose value denotes the node's capacity. Stages 1, 2 and 3 have a total capacity of hosting 2, 3 and 4 microbatch flows. Stage 1 is therefore the bottleneck of the decentralized training. The respective cost of the two flows are 6 (top, in blue) and 10 (bottom, in purple). \algo{} inserts a joining node of capacity 5 into stage 1. Following the addition of this new node, the first stage's capacity increases to 7, and the throughput of stage 2 increases to 3, thus now stage 2 becoming the bottleneck and new joining nodes will be added to it.
}

\subsection{Building Pipelines using Flow Requests}

\algo{} uses three main subprocedures to solve decentralized optimization: Request Flow, Request Change, and Request Redirect. Request Flow builds execution pipelines, while the other two reduce training costs once flows are established. These rely only on local knowledge of system membership and node \emph{inflow}/\emph{outflow}. 
\hide{
The \emph{outflow} of a node is all the flow it sends to the subsequent stage. The \emph{inflow} is the flow it receives from a previous stage.
}
In stable conditions, a relay node’s inflow should equal its outflow. Before stabilization, some nodes have unpaired inflows (wanting to send) or unpaired outflows (wanting to receive). Data nodes start with unpaired flows matching their capacity. Sending outflow reduces capacity by 1, even if unpaired. Each flow has a unique ID, target data node, and current cost to sink—calculated in reverse, from the last stage to the first.


During an iteration, nodes that have available capacity and are in a stable state (no unpaired inflow or outflow) look for a node in a subsequent stage with an unpaired outflow to a specific data node. If multiple such nodes exist, a node $i$ prefers the node $j$ that minimizes (the sum of the cost of flow from $j$ to the sink + the cost of sending from $i$ to $j$), as per \ref{eq:t}. 
Similarly, nodes with unpaired inflow look for a node that has unpaired outflow with the same data node, as the data node of the unpaired inflow. Nodes with unpaired outflow do not perform this search, but merely respond to requests. 

\textbf{Request Flow.} Once a node has found a suitable node with unpaired outflow, it sends it a Request Flow message with the data node and the cost of the flow of the unpaired outflow it is requesting.
When a node receives a Request Flow, it checks the associated data node and cost coming with the request. If it does have unpaired outflow to that data node at that cost, it approves it and adds inflow which connects to the unpaired outflow. If it does not, it rejects it and informs the requester of its current cost to that destination (which is infinite if it has not unpaired outflow to it). Upon seeing its flow request approved, a node adds it to its unpaired outflow (unless it had already unpaired inflow it can connect it to). It then calculates its minimum cost to that sink as the cost between the two nodes, as explained in Equation~\ref{eq:t}, plus the cost of the flow requested as reported by the other node, and broadcasts it to nodes in previous stages. Nodes that have outflow equal to their capacity do not generate Request Flow calls. 
If a node has no peers it can request flow from, then it aims at minimizing its sending cost to the next stage by communicating with the nodes from its stage. This minimization relies on simulate annealing. This is done by querying all of its same stage peers about their current flows, and checking two conditions. 

\textbf{Request Change.} If two nodes have flow to the same sink but with different next stage peers, one of these nodes can determine if switching these flows would reduce the objective function (e.g., maximum sending cost), and consequently request a switch. For example, Node 1 is communicating with Node 2 to Data Node 9, at an intermediate cost \(d(1,2) = 3\). Node 3 is communicating with Node 4 to Data Node at an intermediate cost \(d(3,4) = 8\). However, if instead Node 1 communicates with Node 4 and Node 3 with Node 2, they would have costs \(d(1,4) = 6\) and \(d(3,2) = 6\). The maximum cost on a flow would thus be minimized (8 to 6). If the second node agrees to the switch, because it has that flow and it also thinks that the objective function will be reduced, then each node now sends its outflow to the next peer of the other one. This is performed through an accept message to the same stage peer and a message to the next stage peer that it has a new incoming peer for the same outflow. Otherwise, if the node doesn't agree, nodes keep their outflows as they are. 

\textbf{Request Redirect.} If a node has capacity and sees a peer's flow between two stages, it checks whether rerouting through itself reduces cost. If so, it sends a Request Redirect. For example, if Node 1 → Node 2 → Node 3 costs 11, but Node 1 → Node 4 → Node 3 costs 9, Node 4 can request the redirect. The original node (e.g., Node 2) checks and, if approved, swaps its outflow and cancels its previous flow.
To avoid local minima, we use simulated annealing~\cite{simulatedannealing}: changes that increase cost may still be accepted with probability $e^{ (cost_{current} - cost_{new}) / T} > U(0,1)$, where $T$ is temperature, reduced after each accepted change by a factor \(\alpha\). This allows occasional costly moves to find better solutions.

Once a system has entered a steady state for a few iterations of the algorithm above, the training can proceed. Since the data transferred is small messages, convergence of this algorithm to a close to optimal steady state execution is significantly faster than a training iteration.

\subsection{Tolerating Crashes}
\label{fault-rec}

After completing a batch, a node sends a COMPLETE message with the batch ID to its upstream peer, allowing latency estimation. During training, if a node does not receive a COMPLETE reply within a set time, it assumes the peer is unresponsive and reroutes traffic using the previous algorithm. If no alternate peer is available, it sends a DENY message upstream, prompting flow redistribution. This process can continue recursively up to the source, which may defer the batch to the next iteration. Nodes that send DENY messages are excluded until they free memory.


\textbf{Crashes during the forward pass.} Crash recovery during the forward pass of relay nodes is done as in SWARM~\cite{swarm}. Pipelines are constructed on the fly and a microbatch is routed through them independently by each peer. Instead of using SWARM's stochastic wiring algorithm, \algo{} uses a flow algorithm
that optimizes sending costs and takes into account the memory constraints of each node. 

\hide{
\begin{figure}[t]
  \centering
  \includegraphics[width = \columnwidth]{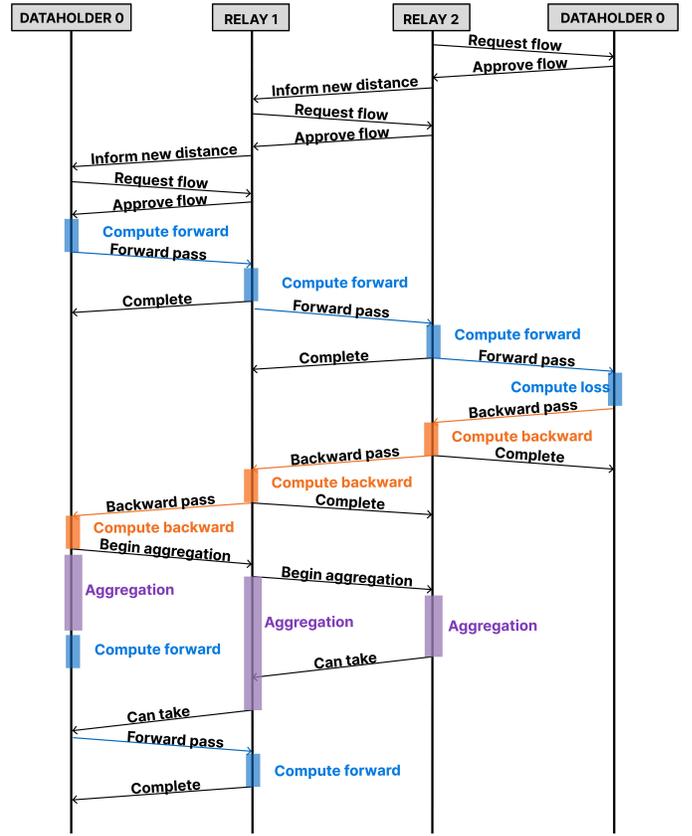}
  \caption{Communication and training happening during one iteration in a simple 3-stage pipeline without any crashes  and with 1 microbatch per iteration. For ease of visualisation messages that relate to the aggregation phase are omitted. 
  }
  \label{fig:messages}
\end{figure}
}

\textbf{Crashes during the backward pass.} 
Nodes send a COMPLETE message after finishing computation. If a node does not receive a COMPLETE message, it notifies the data node. If the data node has no pending microbatches, the pipeline needs not be restored. Otherwise, it pings the first node on the microbatch path. Nodes ping downstream peers along this path, and if a ping fails, the node forwards its activation to a new downstream peer. That peer processes the activation and tries to send results to a previously active node. If unsuccessful, it keeps rerouting until reaching a live node or the data node. The backward pass then resumes from the stored gradient, avoiding recomputation. This recovery is far cheaper than rebuilding the pipeline from scratch, as done in prior work~\cite{swarm}.


\subsection{Training-Aggregation Synchronization}
\label{sec:sync}

Once a new node joins the training and downloads the weights of the stage it will serve, it begins running \algo's flow algorithm
in parallel to the actual training. 
\hide{When a fault occurs it runs one of the two procedures of Section~\ref{fault-rec}. }
Nodes also alternate between training and aggregation phases, which has not been described in previous works~\cite{swarm}. This synchronization is necessary as nodes in the same stage need to have identical parameters when processing microbatches in a iteration. \algo{} relies on a simple algorithm through which nodes signal to each other when this transition happens.
%
%
The aggregation phase starts when the data node leader sends a BEGIN AGGREGATION message, which propagates through the network. Upon receiving it, nodes broadcast and collect model weights within their stage. Once a node completes aggregation and detects that a downstream peer has also finished, it sends a CAN TAKE message upstream to signal readiness for new microbatches. Nodes in the last stage send this without waiting. The system involves several passes: pipeline formation (back to front), forward pass (front to back), backward pass (back to front), relaying of BEGIN AGGREGATION (front to back), and CAN TAKE messages (back to front), marking the end of aggregation and start of a new iteration.
\hide {
This communication pattern is illustrated in Figure~\ref{fig:messages}.
}

\begin{table*}
  \small
  \caption{Performance with crash-prone devices. The average results over 25 repetitions are reported with the standard deviation.}
  \label{table:resultsmain}
  \begin{center}
  \resizebox{\textwidth}{!}{
    \begin{tabular}[H]{p{4.8cm} | p{1.7cm} p{1.7cm} | p{1.7cm} p{1.7cm} | p{1.9cm} p{1.7cm}}
     & SWARM & \algo{} & SWARM & \algo{} & SWARM & \algo{}  \\
     \hline
            
      & \multicolumn{2}{c}{Homogeneous 0\%} & \multicolumn{2}{|c}{Homogeneous 10\%} & \multicolumn{2}{|c}{Homogeneous 20\%}                                          \\
\hline
Time per microbatch (min)            & $\mathbf{0.53\pm0.13} $                           & $0.58\pm0.16$                             & $1.26\pm0.87 $      & $\mathbf{1.01\pm0.37}        $& $1.76\pm1.3$    & $\mathbf{1.17\pm0.43} $\\
Throughput (\#microb/iteration) & $\mathbf{8.0\pm0.13}  $                           & $7.16\pm0.17 $                            & $4.64\pm0.87 $      & $\mathbf{5.8\pm0.37}         $& $\mathbf{6.1\pm1.3}$     & $5.72\pm0.43 $\\

Communication time                   & $6.07\pm4.92 $                           & $\mathbf{4.2\pm2.52}  $                            & $12.23\pm8.81$      & $\mathbf{7.76\pm2.24}        $& $17.74\pm13.15$ & $\mathbf{4.57\pm2.68} $\\
Wasted GPU time                      & $0.27\pm0.0  $                           & $\mathbf{0.03\pm0.0}  $                            & $0.75\pm0.0  $      & $\mathbf{0.2\pm0.0}          $& $1.75\pm0.0  $  & $\mathbf{0.0\pm0.0}   $\\
\hline
& \multicolumn{2}{c}{Heterogeneous 0\%} & \multicolumn{2}{|c}{Heterogeneous 10\%} & \multicolumn{2}{|c}{Heterogeneous 20\%}                                          \\
      
\hline

      Time per microbatch (min)            & $1.59\pm0.21$                            & $\mathbf{1.15\pm0.44}$                             & $4.53\pm4.08$                            & $\mathbf{2.45\pm0.93}$    & $5.36\pm3.56$    & $\mathbf{3.47\pm2.06}$ \\
      Throughput (\#microb/iteration) & $2.96\pm0.21$                            & $\mathbf{3.6\pm0.44}$                             & $1.56\pm4.08$                             & $\mathbf{2.08\pm0.93}$     & $1.72\pm3.56$    & $\mathbf{2.12\pm2.06}$ \\
     
      Communication time                   & $10.55\pm2.79$                            & $\mathbf{3.65\pm1.19}$                           & $3.95\pm1.8$                           & $\mathbf{2.62\pm1.27}$       & $7.18\pm7.22$ & $\mathbf{4.28\pm2.85}$ \\
      Wasted GPU time                      & $0.57\pm0.0$                             & $\mathbf{0.0\pm0.0}$                              & $0.33\pm0.0$                              &  $\mathbf{0.1\pm0.0}$      & $0.33\pm0.0$    & $\mathbf{0.03\pm0.0}$   \\
      \hline

    \end{tabular}
    }
  \end{center}
\end{table*}

\begin{table*}[t]

  \caption{Performance with crash-prone devices for a GPT-like model. The average results over 25 repetitions are reported with the standard deviation.}
  \label{table:resultsnew}
  \begin{center}
  
    \begin{tabular}[H]{p{4cm} | p{1.7cm} p{1.7cm} | p{1.7cm} p{1.7cm} | p{1.7cm} p{1.7cm}}
      & SWARM & \algo{} & SWARM & \algo{}  & SWARM  & \algo{}    \\
      \hline
      & \multicolumn{2}{c}{Homogeneous 0\%} & \multicolumn{2}{|c}{Homogeneous 10\%} & \multicolumn{2}{|c}{Homogeneous 20\%}                                          \\
\hline
Time per microbatch (min)            & $0.68\pm0.08 $                           & $\mathbf{0.37\pm0.01}$                             & $1.77\pm1.25 $      & $\mathbf{0.99\pm0.33}        $& $1.92\pm1.32$    & $\mathbf{1.32\pm0.56} $\\
Throughput & $\mathbf{8.0\pm0.0}  $                           & $\mathbf{8.0\pm0.0 }$                            & $4.52\pm1.35 $      & $\mathbf{6.39\pm1.57}         $& $4.35\pm1.13$     & $\mathbf{6.04\pm0.43} $\\

\hline
& \multicolumn{2}{c}{Heterogeneous 0\%} & \multicolumn{2}{|c}{Heterogeneous 10\%} & \multicolumn{2}{|c}{Heterogeneous 20\%}                                          \\
      
\hline

      \hline
      Time per microbatch (min)            & $1.4\pm0.16$                            & $\mathbf{1.18\pm0.04}$                             & $3.5\pm2.32$                            & $\mathbf{2.63\pm1.08}$    & $5.3\pm4.99$    & $\mathbf{3.16\pm1.79}$ \\
      Throughput & $2.96\pm0.17$                            & $\mathbf{4\pm0.04}$                             & $1.78\pm1.4$                             & $\mathbf{2.26\pm0.93}$     & $1.83\pm3.56$    & $\mathbf{2.39\pm2.06}$ \\
      
      \hline

    \end{tabular}
     
  \end{center}
 
\end{table*}

\section{Performance Evaluation}
\label{sec:eval}

\hide{
In this section, we demonstrate that \algo{} can provide significant speed up for training a large model in decentralized settings, even in the presence of node crashes. 
We further compare the performance of our scheduler against the optimal communication scheduler of Yuan et al.'s DT-FM~\cite{dectrain}. We finally verify the applicability of our system in a practical scenario by showing that the model converges at a rate similar to the one of a centralized solution. 
}

\textbf{Setup.}
Our experiments were performed on LLama and GPT architecture models with varying model sizes. 
We use a private cluster of 5 NVIDIA RTX A4000 GPUs with 16\,GB of GPU memory. Each GPU hosts a number of logical geo-distributed nodes. We simulate the geo-distributed locations by limiting the bandwidth and increasing the latency between logical nodes, with a maximum bandwidth between two nodes in simulated different locations of 500Mb/s and a minimum of 50 Mb/s.
Unless stated otherwise, our decentralized optimization uses $T = 1.7$ for the initial temperature, $\alpha = 0.95$ for the cooling factor.

\textbf{Node Crashes.}
\label{sec:main-results}
We evaluate \algo on a LLaMa-like model (\(d_{model} = 1024\), \(n_{heads} = 18\) and 16 layers) with microbatches of size 4 and sequence length 512. To simulate realistic communication costs, bandwidth is reduced by a factor 32, mimicking activations 32 times larger. Experiments use 18 nodes, with the model split across 6 stages: 3 blocks per stage, except the first, which includes the embedding, 1 block, data retrieval, and loss computation. Two persistent data nodes each push 4 microbatches per iteration.
We compare against SWARM~\cite{swarm}, a crash-tolerant decentralized training method that restarts pipelines after backward pass timeouts. Experiments vary by node capacity and join/leave probability. In the heterogeneous setting, relay node capacities range from 1–3; in the homogeneous case, all are set to 4. Join-leave chance varies from 0\% (no churn) to 10\% (nodes may randomly crash or rejoin each iteration).

We report in Table~\ref{table:resultsmain} the following metrics, averaged over 25 iterations: (1) Minutes per microbatch: The maximum time per iteration (from the slowest data-holder’s perspective), divided by the number of microbatches processed; (2) Throughput per iteration: Total number of microbatches successfully processed in one iteration; and (3) Wasted GPU time: Total pipeline compute time (in minutes) spent on microbatches that were either excluded from aggregation or not part of the final path.
\algo outperforms SWARM in all heterogeneous settings, with up to 45\% speedup under 10\% crash rates. In homogeneous settings, \algo matches SWARM in fault-free cases and outperforms it when crashes occur. SWARM wastes more GPU time due to misrouted microbatches or full pipeline recomputation after faults


We show that GWTF is model-agnostic by repeating tests on a GPT-like model (\(d_{model} = 1024\), \(n_{heads} = 18\) and 16 layers). Results are similar to those with the LLaMa model, but with over 2 times faster iteration time in the homogeneous 0\% crash case. This gain is likely due to GPT’s higher activation communication overhead.


  

      

    

     
 

\begin{figure}[t]
    \centering
    \includegraphics[width=.9\columnwidth]{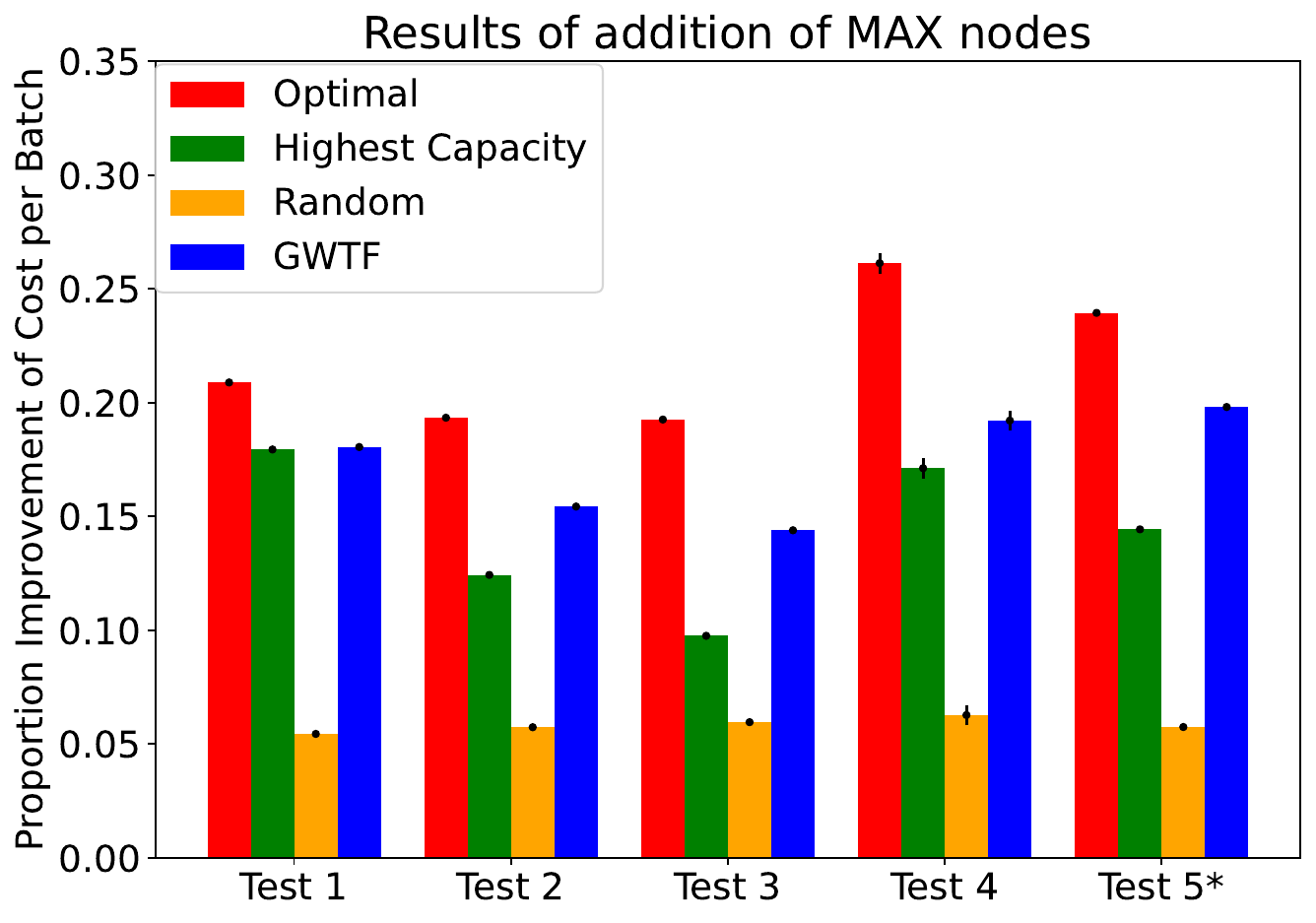}
    \caption{Average improvement over 10 runs with the new node addition tests. Variance is reported with black lines. Higher means better.}
    \label{results:nodeaddition912}
\end{figure}

 \textbf{Handling Joining Nodes.}
We evaluate \algo's handling of joining nodes
depending on the number of stages, the node capacity, and the intra- and interlayer. For each test a total of 97 nodes are used (with 1 of them being a dataholder).
The number of nodes in each stage is the same and can be calculated as \(\frac{N - 1}{S}\) where $N$ is the number of nodes and $S$ the number of stages. The last experiment has a different number of nodes.
In the last of the tests (5*) , the number of nodes per stage is randomly chosen. 
All nodes in the system have properties that adhere to the values described in Table~\ref{table:tests}.
To construct the system, we use the Out-of-kilter algorithm~\cite{outofkilter} to determine the minimum cost flow. 
Iteratively, 20 nodes are added in different stages.
Since every node is a candidate for each stage, we assume that each node knows its costs for each stage. The optimal choice of node addition is determined by running the minimum cost flow algorithm~\cite{outofkilter} for each combination of S candidate nodes added to each of the S stages in the tests. 
Figure~\ref{results:nodeaddition912} reports  the average improvement over 10 runs measured as \(\frac{cost_{now}-cost_{after}}{cost_{now}}\).  
We consider two baselines - adding highest capacity first and adding random nodes. 
%
In all cases \algo{} outperforms the two baselines, however it never achieves an optimal schedule. Still, this optimal schedule cannot be achieved in a decentralised setting - it takes awhile to compute, as it involves trying out every permutation of candidates, and requires global knowledge to run a flow algorithm for every possible permutation. In spite of these limitations, our solution is never more than \(25\%\) slower than the optimal schedule. It is also up to 1.5 times as fast as the highest capacity first baseline and up to 3.5 times faster than the random baseline

\begin{table}
  \caption{Node addition (top) and flow test (bottom) settings. \(\phi\) represents the maximum interlayer cost of a node for the stage it is in/proposing to be in.}
  \label{table:tests}
  \centering
  \begin{minipage}{\columnwidth}
  \resizebox{\columnwidth}{!}{
  \begin{tabular}{lllllll}
    \hline
     & Stages &   Capacities & Interlayer Costs  & Intralayer Costs    \\
    \hline

    1 & 8 &  \(\lfloor U(1,20) \rfloor\) & \(\lfloor U(1,100) \rfloor\)& \(\phi + \lfloor U(50,100) \rfloor\) \\
    2 &  8 &  \(\lfloor U(1,20) \rfloor\) & \(\lfloor U(20,100) \rfloor\)& \(\phi + \lfloor U(50,100) \rfloor\) \\
    
    3 & 8 &  \(\lfloor U(1,5) \rfloor\) & \(\lfloor U(1,100) \rfloor\) & \(\phi + \lfloor U(50,100) \rfloor\)\\
    4 & 12 & \(\lfloor U(1,20) \rfloor\) & \(\lfloor U(1,100) \rfloor\)& \(\phi + \lfloor U(50,100) \rfloor\) \\
    
    5* & 8 &  \(\lfloor U(1,20) \rfloor\) & \(\lfloor U(1,100) \rfloor\)& \(\phi + \lfloor U(50,100) \rfloor\) \\

  \end{tabular}
  }
  \end{minipage}
\begin{minipage}{\columnwidth}
\resizebox{\columnwidth}{!}{
  \begin{tabular}{llllllll}
    \hline
    & Sources & Relays & Stages & Capacities & Link costs   \\
    \hline


    1 & 1 & 40 & 8 & \(\lfloor U(1,3) \rfloor\) & \(\lfloor U(1,20) \rfloor\)\\
    2 & 1 & 40 & 10 & \(\lfloor U(1,3) \rfloor\) & \(\lfloor U(1,20) \rfloor\)\\
    3 & 1 & 40 & 8 & \(\lfloor U(5,15) \rfloor\) & \(\lfloor U(1,20) \rfloor\)\\
    4 & 1 & 40 & 8 & \(\lfloor U(1,3) \rfloor\) & \(\lfloor U(5,100) \rfloor\)\\
    5 & 2 & 40 & 8 & \(\lfloor U(1,3) \rfloor\) & \(\lfloor U(1,20) \rfloor\)\\
    6 & 4 & 80 & 8 & \(\lfloor U(1,3) \rfloor\) & \(\lfloor U(1,20) \rfloor\)\\



  \end{tabular}
  }
  \end{minipage}
\end{table}

\hide{
\begin{table}[t]
  \caption{Flow Test Settings}
  \label{table:flowtests}
  \centering
  \resizebox{.9\columnwidth}{!}{
  \begin{tabular}{llllllll}
    \hline
    & Sources & Relays & Stages & Capacities & Link costs   \\
    \hline


    1 & 1 & 40 & 8 & \(\lfloor U(1,3) \rfloor\) & \(\lfloor U(1,20) \rfloor\)\\
    2 & 1 & 40 & 10 & \(\lfloor U(1,3) \rfloor\) & \(\lfloor U(1,20) \rfloor\)\\
    3 & 1 & 40 & 8 & \(\lfloor U(5,15) \rfloor\) & \(\lfloor U(1,20) \rfloor\)\\
    4 & 1 & 40 & 8 & \(\lfloor U(1,3) \rfloor\) & \(\lfloor U(5,100) \rfloor\)\\
    5 & 2 & 40 & 8 & \(\lfloor U(1,3) \rfloor\) & \(\lfloor U(1,20) \rfloor\)\\
    6 & 4 & 80 & 8 & \(\lfloor U(1,3) \rfloor\) & \(\lfloor U(1,20) \rfloor\)\\



  \end{tabular}
  }
\end{table}
}

 \textbf{Optimality.}
%
%
We compare the end-to-end training time of \algo against a GPipe setting with a communication optimal arrangement, calculated by DT-FM~\cite{dectrain}. The setting of the experiment are mostly identical to those of the prior 0\% homogeneous setting. Following prior work~\cite{dectrain} we use several pipelines with 4 microbatches per pipeline. In order for the two system to be comparable, we have 3 dataholders and 15 relay nodes distributed across 6 stages (3 nodes per stage). 
\hide{
The end-to-end training time between the two systems is compared in Table~\ref{table:resultsoptimal}.
}
Although the optimal computation schedule outperforms \algo by almost \(13\%\) (0.44\,s and 0.51\,s for DT-FM and \algo, respectively), it takes much longer to be computed, as it involves the use of a genetic algorithm~\cite{dectrain} and scales exponentially with the number of nodes. \algo is therefore approaching the optimal schedule and can be used at scale.

\hide{
\begin{table}[t]
\caption{Comparison against optimal schedule}
\label{table:resultsoptimal}
\begin{center}
\begin{tabular}[H]{p{2cm}|p{2.4cm}p{2.4cm}p{1.6cm}}
   & Time per microbatch  & Throughput per iteration    \\ 
  \hline
  DT-FM~\cite{dectrain} & $0.44 \pm 0.0$ & $9.0\pm 0.0$  \\
  \hline
  
  \algo{} & $0.51 \pm 0.08$ & $8.08 \pm 0.08$ \\ 
  \hline

\end{tabular}
\end{center}
\end{table}
}

\begin{figure}[t]
    \centering
    \includegraphics[width=.9\columnwidth]{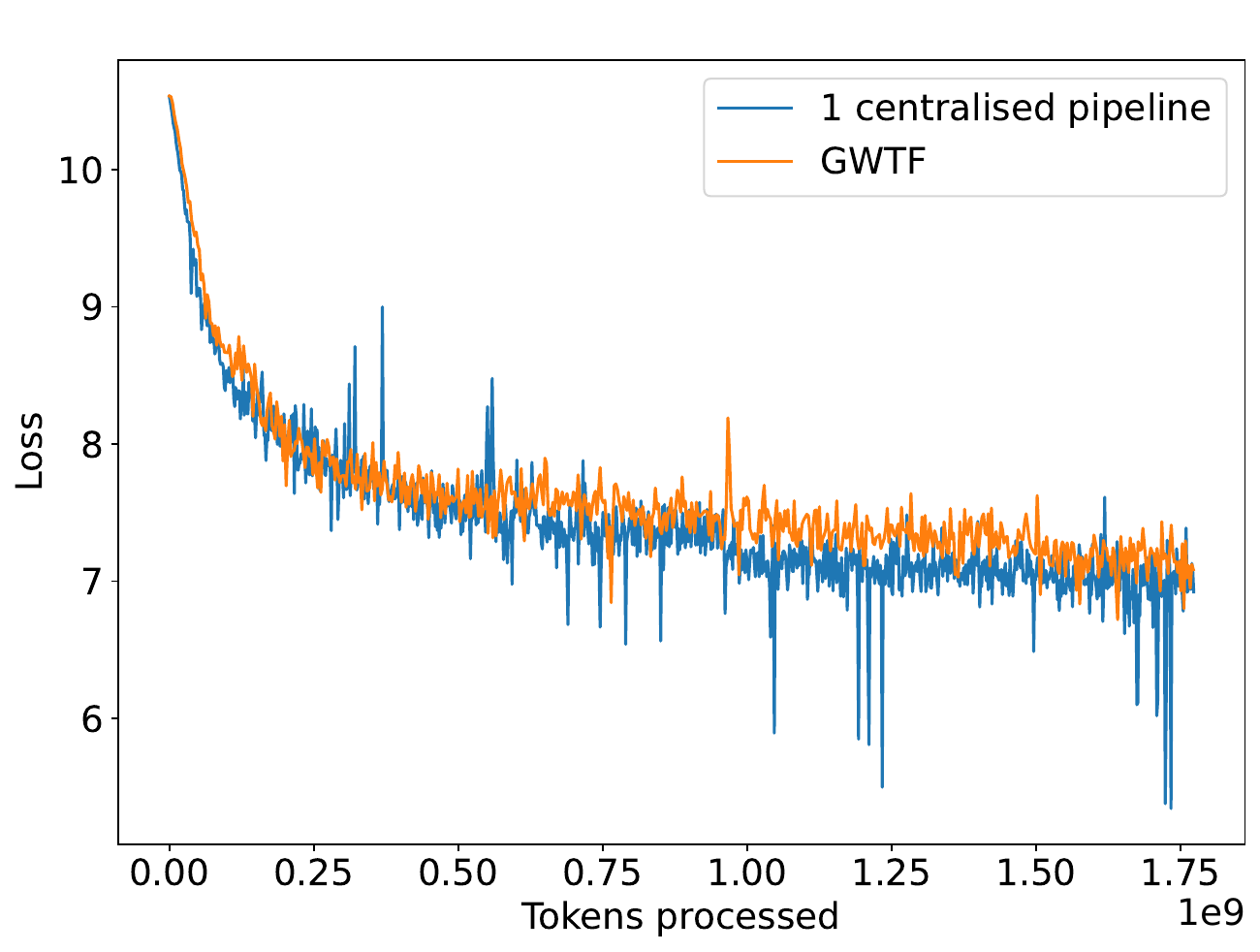}
    \caption{Loss convergence} 
    \label{results:convergence}
\end{figure}

\textbf{Training Convergence.}
\hide{
Since our system does not modify the training of the model (always the entire model is ran as in a centralized solution), }
\algo has 
the same theoretical convergence as SGD. We verify this by training our system with 10\% crash rate and a single GPipe pipeline of 8 nodes, with 8 microbatches of size 1 and sequence length 4096. Our system's hyperparameters were heterogeneous nodes with 10\% crash chance and 1 data node. The model used in this experiment is the LLaMA-7b distributed uniformly across 8 stages. The dataset used was the Wikipedia English dataset~\cite{wikidump}. Figure~\ref{results:convergence} confirms that \algo{} has a similar convergence of loss as a centralized solution with the same batch size.

\textbf{Ablation studies.}
We evaluate our flow algorithm in 6 different settings (cf. Table~\ref{table:tests}). The first four settings involve a single source-sink node while the last two have multiple source-sink nodes. 
\hide{
Details about the tests are presented in Table~\ref{table:flowtests}. 
In all cases the source-sinks were given sufficient capacity to prevent bottlenecks.}
In order to compare to the optimal result of Fulkerson's algorithm~\cite{outofkilter}, our procedure attempts to minimize the sum of the costs of all flows, \(min\left(\sum_{i,j \in \textbf{N}} d(i,j) * f(i,j)\right)\). Tests 5 and 6 are not compared with the optimal baseline, as there the formulation differs, in that a source must deliver to a specific sink. We use the approach from SWARM~\cite{swarm} of sending to the next stage closest node as a baseline. Experiments are evaluated on at most 120 iterations (roughly a minute) of the different algorithms. 
\hide{As Figure~\ref{results:flowtests} shows,} 
\algo consistently outperforms SWARM by up to 50\% in heterogeneous communication settings. 
\hide{
This partially explains the higher throughput our system achieves compared to Swarm~\cite{swarm}, as it greedy approach can choose significantly worse paths. 
}
Additionally, due to the negligible time required to compute the paths relative to the actual training, the cost of using this approach instead of SWARM's greedy approach is outweighed by training time reduction.

\hide{
\begin{figure}[t]
    \centering
    \includegraphics[width=.9\columnwidth]{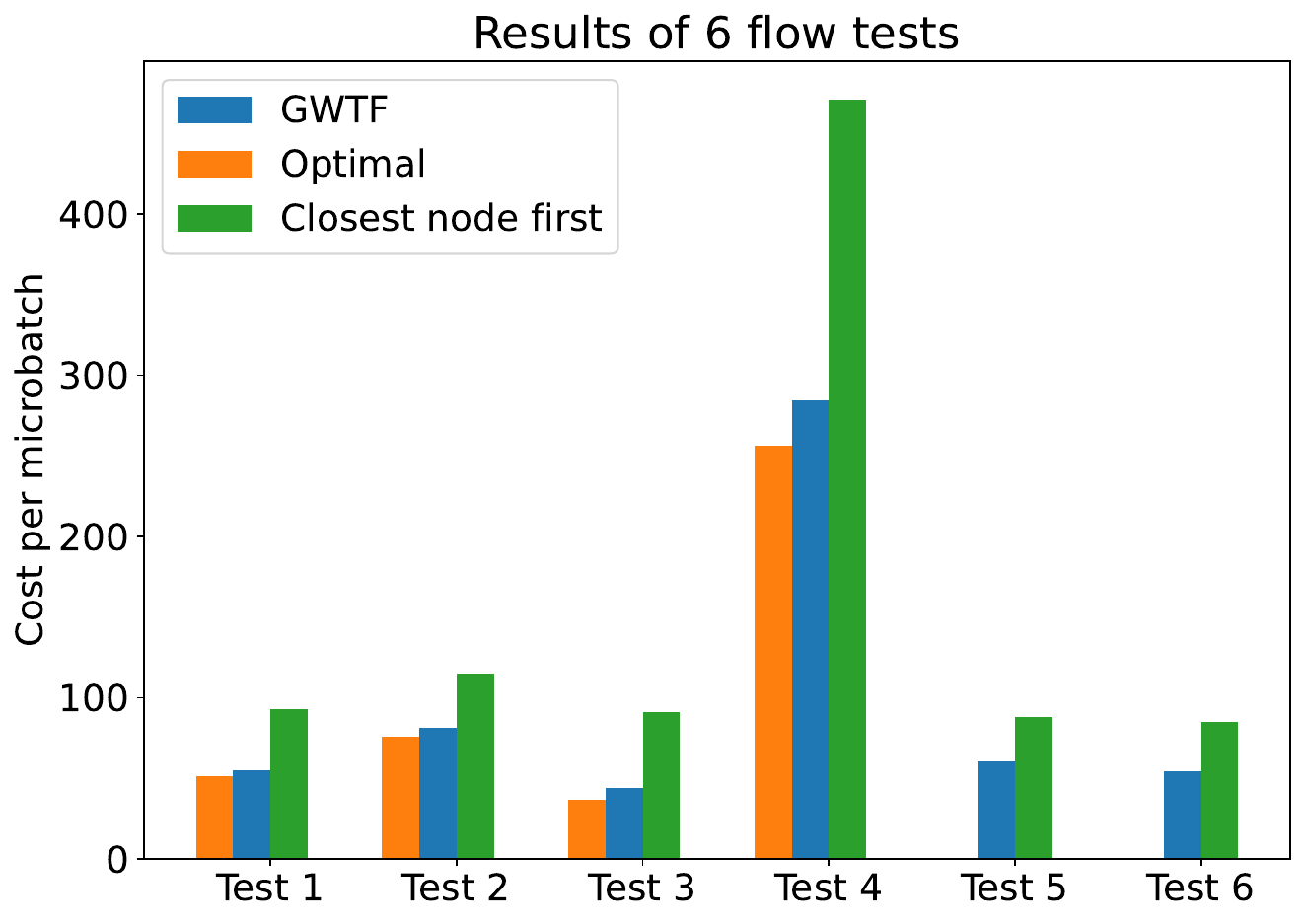}
    \caption{Average cost per microbatch in flow tests}
    \label{results:flowtests}
\end{figure}
}

    
    

\section{Discussion and Future Work}

\paragraph{Byzantine Nodes} While this work does model nodes as crash-prone, it does not consider nodes who might deviate from the protocols described. Such nodes are typically referred to as Byzantine~\cite{byzantine} and could, for example, prevent convergence in this setting by sending random activations. In order for any system do be viable in any practical setting, such nodes need to be dealt with in an efficient manner. Unfortunately, while the literature on training large language modes in a decentralised setting is greatly lacking, work on training them in a setting with Byzantine nodes is non-existent. 

\paragraph{Checkpointing} While this paper assumes that at least one node remains active per stage, in a realistic setting this is not guaranteed. However, such issues are typically addressed via check pointing (storing current parameters on another remote device)~\cite{gemini}. Recent work on this matter assumes a stable, non-faulty, central node, which can store the check points~\cite{gemini,checknrun}. However, this is insufficient for our setting. Efficient decentralised checkpointing with crash-prone devices remains an unexplored topic.

\paragraph{Incentives} Throughout this paper we assume volunteer nodes - nodes sparing some resources for no compensation. Training of large language models can be both computationally and time intensive. As such, many individuals may be discouraged to participate in the training, due to the energy costs they might be faced. However, as seen by the recent popularity of blockchains in the mainstream~\cite{nytimes}, individuals may be incentivised to participate at the promise of a reward proportional to their contribution. Blockchains are especially well suited for this setting, as rewards can be handed instantaneously, can be tied to an existing asset, and offer good scalability with increased number of workers. Recent work has explored this idea in depth and we consider it as a possible extension of this system~\cite{pouai,bc1}
\paragraph{Different Architectures} This work primarily focuses on Large Language Models for its experiments. However, the ideas presented are not LLM exclusive. Any system, which necessitates either pipeline parallelism or data parallelism can make use of parts or the whole of our framework. Vision Transformers share a similar architecture as Large Language Models and also benefit from pipeline parallelism, due to their growing sizes~\cite{vitpp}. Surprisingly, even convolutional models such as ResNet and VGG can benefit from pipeline parallelism (though it requires pipelining of microbatches with some staleness awareness to maximise efficiency)~\cite{pipedream}. For both scenarios, the framework presented in this work can in theory be utilised.

\section{Conclusion}
\label{sec:conclusion}
We described \algo, the first crash tolerant decentralized training framework that aims to minimize training time and maximize its throughput.  \algo{} models the decentralized training procedure as a flow problem, and effectively decides its execution on heterogeneous clients and network links. \algo{} is structured in two key modules: decentralized flow computation and tolerating crashes. We evaluate \algo{} on decentralized training  LLama-like and GPT-like models in a geo-distributed setting, against SWARM and GPipe, on both homogeneous and heterogeneous setups with different node churning dynamics. Our results show that \algo{} is able to improve the training time by up to a 45\% and throughput by up to a 30\% increase while wasting almost zero GPU time of any joining clients.  

\bibliographystyle{IEEEtran}
\bibliography{sample-base}


\end{document}